# Collocation2Text: Controllable Text Generation from Guide Phrases in Russian

**Vychegzhanin S. V.**
Vyatka State University,
Kirov, Russia
vychegzhaninsv@gmail.com

**Kotelnikov E. V.**
Vyatka State University,
Kirov, Russia
kotelnikov.ev@gmail.com

### Abstract

Large pre-trained language models are capable of generating varied and fluent texts. Starting from the prompt, these models generate a narrative that can develop unpredictably. The existing methods of controllable text generation, which guide the narrative in the text in the user-specified direction, require creating a training corpus and an additional time-consuming training procedure. The paper proposes and investigates *Collocation2Text*, a plug-and-play method for automatic controllable text generation in Russian, which does not require fine-tuning. The method is based on two interacting models: the autoregressive language ruGPT-3 model and the autoencoding language ru-RoBERTa model. The idea of the method is to shift the output distribution of the autoregressive model according to the output distribution of the autoencoding model in order to ensure a coherent transition of the narrative in the text towards the guide phrase, which can contain single words or collocations. The autoencoding model, which is able to take into account the left and right contexts of the token, "tells" the autoregressive model which tokens are the most and least logical at the current generation step, increasing or decreasing the probabilities of the corresponding tokens. The experiments on generating news articles using the proposed method showed its effectiveness for automatically generated fluent texts which contain coherent transitions between user-specified phrases.

**Keywords:** text generation; GPT; BERT
**DOI:** 10.28995/2075-7182-2022-21-564-576

# Collocation2Text: Управляемая генерация текста по направляющим выражениям на русском языке

**Вычегжанин С. В.**
Вятский государственный
университет,
Киров, Россия
vychegzhaninsv@gmail.com

**Котельников Е. В.**
Вятский государственный
университет,
Киров, Россия
kotelnikov.ev@gmail.com

### Аннотация

Крупные предобученные языковые модели способны создавать разнообразные и связные тексты. Эти модели, отталкиваясь от затравки, генерируют повествование, которое может уходить в непредсказуемом направлении. Существующие методы управляемой генерации текста, направляющие повествование в тексте в заданном пользователем направлении, требуют создания обучающего корпуса и выполнения трудоемкой процедуры дополнительного обучения. В статье предложен и исследован plug-and-play метод автоматической управляемой генерации текстов на русском языке *Collocation2Text*, не требующий процедуры дообучения (fine-tuning). Метод основан на взаимодействии двух моделей: авторегрессионной языковой модели ruGPT-3 и автокодирующей языковой модели ruRoBERTa. Идея метода заключается в смещении выходного распределения авторегрессионной модели в соответствии с выходным распределением автокодирующей модели для обеспечения связного перехода повествования в тексте к направляющему выражению, которое может содержать одиночные слова или словосочетания. Автокодирующая модель, способная учитывать левый и правый контекст токена, «подсказывает» авторегрессионной модели, какие токены являются наиболее и наименее логичными на текущем шаге генерации, повышая или понижая вероятности соответствующих токенов. Эксперименты по генерации новостных статей с использованием предложенного метода показали его эффективность для автоматического создания гладких текстов, содержащих связные переходы между заданными пользователем выражениями.

**Ключевые слова:** генерация текстов; GPT; BERT

## 1    Introduction

Natural language generation (NLG) is a field of computational linguistics that deals with the construction of computer systems which can generate understandable texts in English or other human languages [8, 29]. At an early stage of development, language generation was carried out using rule-based models, hidden Markov model, as well as shallow models trained on sparse multidimensional features. Due to large neural network language models, in particular, models based on the Transformer architecture [33], trained on dense vector representations, automatically generated texts showed a new level of quality. Modern neural network models such as GPT-2 [28] and GPT-3 [3] can generate impressively realistic human-like texts.

NLG has a wide range of applications, including generation of answers to user questions in dialogue systems (chatbots) [31] and question-answer systems [6], generation of stories [1], poetry [22], news articles [32], product and service reviews [7].

Making text generation controllable is an important fundamental issue in NLG. For example, when generating stories, it is necessary to control the storyline or the completion of the story. Controllable Text Generation (CTG) is the task of generating a natural language text with given attributes [26]. Topic, sentiment, keywords, entities, events, etc. can be considered as attributes.

Soft and hard controls can be applied over text generation models. Soft control aims at ensuring, e.g., the desired sentiment or general topic of the generated text. The aim of hard control is to ensure that some explicit constraints, e.g., specific keywords are contained in the text. Figure 1 shows an example of CTG with hard control when generating a story with keywords from the storyline in a given order.

| Prompt | The student needed a laptop for study. |
|---|---|
| Storyline | store → bought → installed → happy → project |
| Generated text | The student needed a laptop for study. He came to the computer **store**. The student chose and **bought** a new laptop. He **installed** the necessary software. The student is **happy** to do a study **project** using a new laptop. |

Figure 1: Example of controllable story generation with hard control

The problem of most existing models with soft [4, 14] and hard [5, 36] controls lies in the need to create a training corpus and perform a training procedure. Creating such a corpus and additional training of the model is difficult, expensive and time consuming. This paper proposes to overcome this problem by developing a plug-and-play method that can be applied to pre-trained large language models. It should be noted that currently there is a lack of research on controllable text generation in Russian, thus, the proposed method was tested on the Russian language models.

The idea of the method is to shift the output distribution of the autoregressive language model (ruGPT-3) according to the output distribution of the autoencoding language model (ruRoBERTa). The method provides a coherent transition of the narrative in the text to the guide phrase. It is a plug-and-play method, i.e., it can be used with any autoregressive and autoencoding language models that have vocabularies built using the same encoding algorithm. Within our study, the experiments on generating news articles proved the effectiveness of the proposed method for creating a coherent text from the list of guide phrases.

The contributions of this paper are the following:

- we propose a method of controllable text generation *Collocation2Text* that generates texts according to a user-specified list of guide phrases, which can be either single words or collocations;
- we apply this method to the Russian language;
- we conduct experiments with article generation in order to confirm the effectiveness of the proposed method.

## 2    Previous Work

This section discusses the existing methods for text generation in relation to the story generation task, which is of the main interest in our study. A story is a description of real or imaginary characters and events generated to achieve one or more goals, e.g., to entertain or educate [1]. The difficulty of the neural network story generation task consists in generating a coherent and fluent story that is much longer than a short input user-specified prompt. Story generators can be classified into three categories:

– structural models which generate a structured story by dividing the story into slots following a given scheme [12, 16];

– planning-based models which generate a story as a chain of causally connected events to pursue a final goal [9, 13];

– machine learning models which learn the conditional probability distribution between story events from a story corpus [18, 34].

Currently, the greatest success in automatic story generation has been achieved by machine learning models, in particular, deep neural networks. A number of previous studies in this direction should be noted. Jain et al. [11] used a recurrent neural network with an attention mechanism and gated recurrent units. As input, the neural network received standalone textual descriptions describing an event or scene and converted them into coherent summaries. However, according to the results of the experiments, the summaries obtained were not completely semantically related to the input descriptions. To overcome this disadvantage, Fan et al. [5] decomposed story generation into two stages. During the first stage, they generated the story premise representing the structure of the story using the convolutional language model. At the second stage, the sequence-to-sequence model was used to create a story that followed the premise.

Peng et al. [24] developed a framework to control story ending valence (happy or sad ending) and a storyline based on a recurrent neural network generation model with an attention mechanism. The experiments showed that through introducing storylines the coherence of the story is improved compared to uncontrolled generation models.

Yao et al. [36] proposed another hierarchical story generator that combines storyline planning and text generation. During the learning stage, for each story from the corpus, a storyline was built from the most important words in each sentence using the RAKE algorithm [30]. Then, the storyline was converted to a text using the sequence-to-sequence model. It was proved that with explicit storyline planning the generated stories are more diverse and coherent than those generated without creating a full plan.

The examples of texts generated by the GPT-2 model [28] showed that large pre-trained language models are able to generate texts similar to human-written texts. These models have been actively used by researchers to generate stories. Dathathri et al. [4] proposed a plug-and-play method that trains an external attribute discriminant model. The gradients from the attribute model "push" the hidden activation of the pre-trained language model on the Transformer architecture to guide the target text generation.

Pascual et al. [23] proposed another plug-and-play method of controllable text generation, which can be applied to the existing autoregressive language model without additional training. The authors created a strategy for controlled decoding, in which the generated text contains words from the given guide word sequence. The idea of the method is that in the process of generating the next word the output distribution of the language generation model is shifted to the semantic space of the guide word. The degree of similarity between the tokens of the language model and the guide word is calculated as the cosine measure in the vector space word2vec [21] or GloVe [25]. The method not only encourages the guide word to occur explicitly, but also encourages the model to generate the appropriate context for the guide word to occur.

Recently, significant progress in the control of autoregressive models has been made using prompt-tuning. For example, Li and Liang [17] proposed prefix-tuning, which keeps language model parameters frozen but optimizes a small continuous task-specific vector. Qin and Eisner [27] proposed to learn a mixture of soft prompts. Liu et al. [19] designed an optimized and adapted implementation of deep prompt tuning for generation.

Our study is consistent with [23]. The difference of our method is that the output distribution of the autoregressive language generation model is not shifted to the guide word in the word2vec or GloVe vector space, but is summed with the output distribution of another autoencoding language model. In addition, the proposed method allows using single words and collocations in the guide sequence.

### 3    A method of controllable text generation

The proposed *Collocation2Text* method can be applied to any autoregressive language model for which the probability of a sequence of linguistic units (characters, tokens, words, sentences) $X = \{x_1, \dots, x_n\}$ is decomposed using the chain rule:

$$p(X) = \prod_{i=1}^{n} p(x_i | x_1, x_2, \dots, x_{i-1}) = \prod_{i=1}^{n} p(x_i | X_{<i}). \tag{1}$$

We consider models $p$ that assign a probability to all sequences $X$ in the space of strings $Y(V, n_{max}) = V^{n_{max}}$, where $V$ is the model vocabulary and $n_{max}$ is the maximum sequence length.

The task of text generation based on the autoregressive language model is to decode sequences of linguistic units from the distribution $p$. In the generation process, a score function and a decoding algorithm are of the most importance. The score function assigns some degree of significance for generating the resulting text to each sequence of linguistic units. The score function depends on the sequence probability and can be modified according to the requirements for the generation method. Formally, the score function of the autoregressive language model can be defined as a map from strings generated under the model vocabulary to a real number and written in the form $score_{alm}(\cdot | X_{<i}) : V^i \to \mathbb{R}$. For such models, the default score function is usually the log-probability:

$$score_{alm}(\cdot | X_{<i}) = \log p(\cdot | X_{<i}). \tag{2}$$

A decoding algorithm is a class of algorithms that decode a text according to the score function. Examples of such algorithms are beam search [20] and nucleus sampling [10]. We apply top-$K$ sampling, which consists in choosing a token from the $K$ tokens with the largest values of the score function.

Autoregressive models (e.g., GPT-3) are trained to predict the next token in a sequence by the previous context. For such models, text generation is a natural application. However, their disadvantage in the task of controllable text generation according to given guide phrases is that in the process of predicting the next token the context following this token is not taken into account. In contrast, autoencoding (masked) language models (e.g., BERT) are trained to predict masked tokens in the text and reconstruct the original text. Such models take into account the left and right contexts of the predicted token. We combine the advantages of the two classes of models as follows. The autoencoding language model, given the left and right contexts of the token, helps the autoregressive model change the output distribution, increasing the probability of more suitable tokens and decreasing the probability of less suitable tokens for the coherent transition to the guide phrases.

In this paper, we propose the *Collocation2Text* method, which modifies the autoregressive model score function (2) by shifting its values based on the output distribution of the autoencoding model. We consider a probabilistic language generator $p$, a guide phrase $W = \{w_1, \dots, w_m\}$ which should occur in the generated text and a sequence of tokens $X = \{x_1, \dots, x_{i-1}, x_i\}$. We propose to modify the score function for the current token $x_i$ as follows:

$$score'_{alm}(x_i | X_{<i}, W) = score_{alm}(x_i | X_{<i}) + score_{mlm}(x_i | X_{<i} \cup W), \tag{3}$$

where $score_{mlm}(x_i | X_{<i} \cup W)$ is a score function for the masked language model, $X_{<i} \cup W = \{x_1, \dots, x_{i-1}, w_1, \dots, w_m\}$.

In equation (3), the output score is calculated as the sum of the score of the autoregressive model based on the left context of token $x_i$ ($X_{<i}$) and the score of the autoencoding model based on the left ($X_{<i}$) and right ($W$) contexts of token $x_i$.

In order to increase the probability of the first token $w_1$ from the guide phrase $W$ in the text, the value of the score function for $w_1$ is modified as follows

$$score'_{alm}(w_1 | X_{<i}, W) = score'_{alm}(x_K | X_{<i}, W) + \lambda_i \cdot \alpha_i \cdot \Delta_i, \tag{4}$$

where $\lambda_i$ is a hyperparameter indicating the strength of the score shift of the token $w_1 \in W$ at the $i$-th generation step, $\alpha_i$ and $\Delta_i$ are parameters determined by equations:

$$\alpha_i = \frac{s_{w1} - s_{min}}{s_{max} - s_{min}}, \tag{5}$$

$$\Delta_i = s_{max} - s_K, \tag{6}$$

where $s_{w1} = score'_{alm}(w_1|X_{<i}, W)$ is the value of the score function for the first token $w_1$ of the guide phrase $W$; $s_{min} = \min\big(score'_{alm}(x_i|X_{<i}, W)\big)$ is the minimum value of the score function; $s_{max} = \max\big(score'_{alm}(x_i|X_{<i}, W)\big)$ is the maximum value of the score function; $s_K = score'_{alm}(x_K|X_{<i})$ is the value of the score function for the last token on the list of top-$K$ tokens in the top-$K$ sampling decoding strategy.

Thus, we put the token $w_1$ in the list of top-$K$ tokens. The position of $w_1$ on the top-$K$ list depends on the initial relative position $\alpha_i$ of this token.

Figure 2 shows a scheme of increasing the probability of the first token $w_1$ from the guide phrase $W$ at the $i$-th step of the top-$K$ sampling decoding strategy according to equation (4).

The coefficient $\lambda_i$ determines the speed of the appearance of the guide phrase in the text. As $\lambda_i$ increases, the first token of this phrase rises in top-$K$. The value of the coefficient $\lambda_i$ increases at each $i$-th step of the generation process according to the equation:

$$\lambda_i = \lambda_0(i - i_n), \tag{7}$$

where $\lambda_0$ is the initial value of the shift; $i_n$ is the number of the step at which the previous guide phrase appeared.

The proposed method ensures the interaction between neural network models of two types. The autoencoding model, which is able to take into account the left and right contexts of the token, "tells" the autoregressive model which tokens are the most logical at the current decoding step, shifting the values of the score function according to equation (3). Additionally, at each decoding step, the value of the score function for the first token of the guide phrase $W$ is increased and thus the probability of choosing this token is increased. After occurring in the text, the token $w_1$ is replaced by the whole phrase $W$.

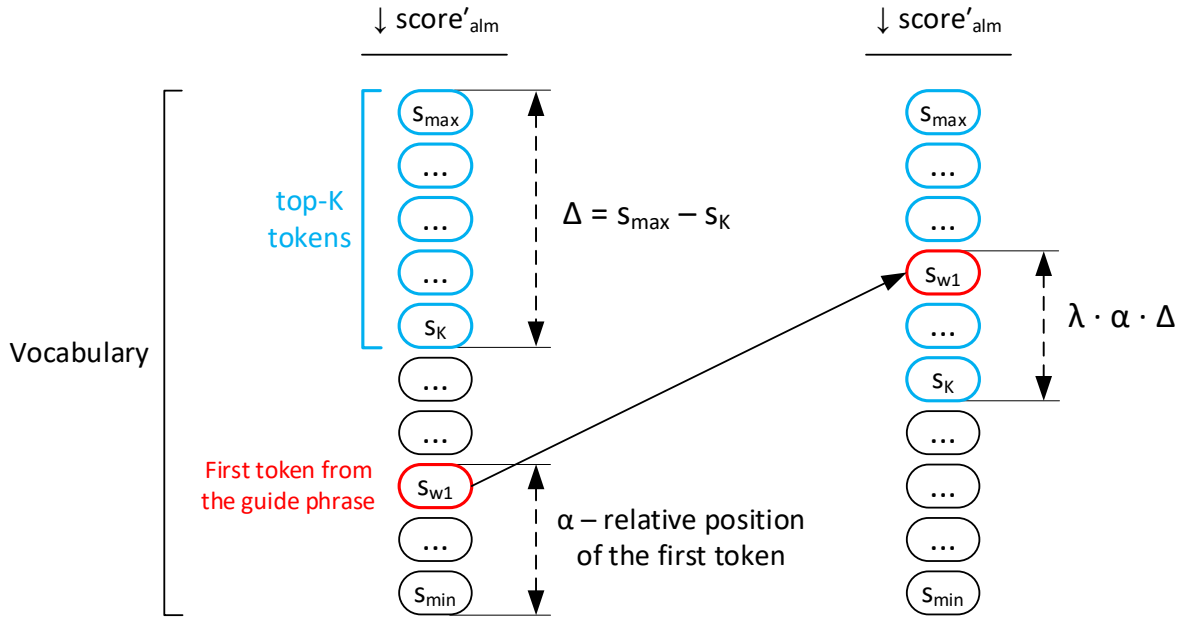

Figure 2: Scheme for increasing the probability of the first token
of the guide phrase according to equation (4)

To illustrate the proposed method, we consider a text at an $i$-th generation step and a guide phrase, separated by a special <mask> token (Figure 3).

---

Мэр рассказал о планах благоустройства городских парков. Сегодня в городе <mask> очередная пресс-конференция

The mayor told about the plans for the improvement of city parks. Today in the city <mask> another press conference

---

Figure 3: Text example at the *i*-th generation step (Russian and English versions)

In Figure 3, the sentence used as a prompt for the autoregressive model is highlighted in blue, the guide phrase is highlighted in orange. By the *i*-th generation step, the autoregressive model has already generated the phrase "Сегодня в городе" ("Today in the city"). At the *i*-th step, as input, the model receives the text located to the left of the <mask> token and returns the distribution $score_{alm}$ over the tokens of its vocabulary. The autoencoding model takes the whole text as input, including the <mask> token and the guide phrase, returning the distribution $score_{mlm}$ over the tokens of its vocabulary. Then the distributions returned by the models are summed according to equation (3), and the value of the score function for the first token of the guide phrase is also increased according to equation (4). The resulting output distribution is subjected to the top-*K* sampling decoding algorithm that selects a token from the model vocabulary to be placed instead of the <mask> token. In the next step, the <mask> token moves one position to the right. The guide phrase is inserted until it occurs in the text.

## 4    Experimental Setup

In the experiments, ruGPT3Large[1] (760 million parameters) was used as an autoregressive language model, which is a Russian-language adaptation of the GPT-2 model [28], and ruRoBERTa-large[2] (355 million parameters) was used as an autoencoding language model.

To check whether the word generated by the ruGPT3Large model is on the list of guide phrases, we compared the lemma of this word and the lemma of the first word of the guide phrase. The morphological analysis was performed using the *pymorphy2* library[15].

In the experiments, news articles were generated according to a storyline containing from three to six guide phrases, which were single words or collocations. We chose the news domain since methods for generating news articles are widely used in computer journalism, e.g., to provide journalists with first draft news and templates [2]. In addition, the study of text generation methods in relation to news is an important direction of NLG because it allows evaluating the ability of language models to create disinformation [32]. Table 1 shows ten prompts with guide phrases used in the experiments.

The following news generation strategies were compared:

- uncontrolled language generation, i.e., using only the base ruGPT3Large model (ruGPT3 strategy);
- shifting the token scores according to equation (3) without increasing the probability of choosing the first token of the phrase $W$ by the model $p$ (ruGPT3 + ruRoBERTa strategy);
- shifting the token scores according to the equation (3) with increasing the probability of choosing the first token of the phrase $W$ according to equation (4) (ruGPT3 + ruRoBERTa + $\lambda\alpha\Delta$ strategy).

Top-*K* sampling with the parameter $K = 10$ was used for decoding. The length of the generated sequence was 90 tokens (excluding the prompt).

The quality of the generated texts was evaluated using the measures of perplexity, repetition, and success rate [4, 23]. Perplexity (PPL) is calculated as exponential average of the negative logarithmic probability per token in the language model. A separate ruGPT3Medium[3] model (350 million parame-

---

ters) was used to compute the perplexity. Repetition score (Rep) [35] calculates the proportion of repeated 4-grams in the text. Success rate (SR) calculates the proportion of guide phrases which occurred in the generated text.

## 5 Results and Discussion

The vocabularies of the ruGPT3 and ruRoBERTa models differ from each other: the size of both is 50,257 tokens, and the intersection of these vocabularies contains 35,215 tokens. Thus, the proportion of common tokens in model vocabularies is 70.1%. Both models use BBPE tokenization (Byte-level Byte-Pair-Encoding) as the tokenization strategy. The differences in the vocabularies are due to the differences in the training corpora used to pretrain the models. The training corpus for the ruGPT3 had a total size of over 600 Gb. It included a huge collection of Russian literature, Russian and English Wikipedia, public sections of Pikabu[4], a complete collection of materials from the popular science portal 22century.ru and the banking portal banki.ru, as well as the Omnia Russica corpus[5]. The training corpus for ruRoBERTa had a size of 250 Gb. It was created from the ruGPT3 training corpus, from which the English texts were removed. Examples of ruGPT3 and ruRoBERTa tokenization are shown in Figure 4.

---

Text: В этом году конкурс в вузы на ИТ-направления сильно вырос.

ruGPT3: ['В', ' этом', ' году', ' конкурс', ' в', ' ву', 'зы', ' на', ' ИТ', '-', 'на', 'прав', 'ления', ' сильно', ' вырос', '.']

ruRoBERTa: ['В', ' этом', ' году', ' конкурс', ' в', ' в', 'узы', ' на', ' И', 'Т', '-', 'направ', 'ления', ' сильно', ' вырос', '.']

---

Figure 4: An example of ruGPT3 and ruRoBERTa tokenization

Table 2 presents the average values of perplexity, repetition, and success rate, calculated for news articles generated from 10 prompts and guide phrases presented in Table 1. We generated 10 samples of news per prompt. Thus, the average values of the measures for each strategy were calculated using 100 samples.

In Table 2, the values of SR show that the ruGPT3 + ruRoBERTa + $\lambda\alpha\Delta$ strategy ensures the occurrence of more than 96% of guide phrases in the text, while in the texts generated by the ruGPT3-such phrases are absent. When using the ruGPT3 + ruRoBERTa strategy, the guide phrases occurred in relatively few texts, as indicated by the value SR = 13.63%.

Analyzing the Rep values, we can conclude that the strongest Rep = 61.35% when using the ruGPT3 + ruRoBERTa strategy. The cyclic repetition of the same fragments was observed in all the texts. The ruGPT3 + ruRoBERTa + $\lambda\alpha\Delta$ strategy has the least repetition, which does not exceed 4%.

---

[4] https://pikabu.ru.
[5] https://omnia-russica.github.io.

| # | Prompt | Guide phrases |
|---|--------|---------------|
| 1 | В пятницу в драматическом театре состоялась премьера спектакля.<br><br>On Friday, the play premiered at the Drama Theatre. | новое прочтение, игра актеров, красивые декорации, реакция зрителей, успех<br><br>new reading, play of the actors, beautiful scenery, reaction of the audience, success |
| 2 | В пятницу в драматическом театре состоялась премьера спектакля.<br><br>On Friday, the play premiered at the Drama Theatre. | произведение классической литературы, ошибка режиссера, освещение, провал спектакля<br><br>work of classical literature, director's mistake, lighting, performance failure |
| 3 | Все билеты на очередной этап гонок «Формула-1» были раскуплены задолго до соревнования.<br><br>All tickets for the next racing event of the Formula 1 had been sold out long before the competition began. | опытные гонщики, мощные двигатели, визг тормозов, рев толпы, восторг зрителей<br><br>experienced racers, powerful engines, screeching brakes, roaring crowd, delighted spectators |
| 4 | В этом году конкурс в вузы на ИТ-направления сильно вырос.<br><br>This year, the acceptance rate for IT specialities in universities has grown significantly. | количество выпускников школ увеличилось, интересная специальность, перспективы хорошей работы, высокая зарплата<br><br>the number of school graduates has increased, interesting speciality, good job prospects, high salary |
| 5 | Стоимость автомобилей в автосалонах выросла на 25% с начала года.<br><br>The cost of cars in showrooms has increased by 25% since the beginning of the year. | новые автомобили, средняя комплектация, долгое ожидание, отсутствие комплектующих<br><br>new cars, medium equipment, long wait, lack of components |
| 6 | После снятия ограничительных мер посещаемость кинотеатров увеличилась.<br><br>After lifting the restrictive measures, cinema attendance increased. | сильная эпидемия, длительный карантин, рост на 10%<br><br>severe epidemic, long quarantine, 10% growth |
| 7 | Родители и школьники дают разную оценку дистанционному образованию.<br><br>Parents and schoolchildren assess distance education differently. | вспышка коронавируса, обучение в дистанционном формате, родители высказали недовольство, ученики стали позже вставать<br><br>outbreak of coronavirus, distance learning, parents expressed dissatisfaction, schoolchildren get up later |
| 8 | Мэр рассказал о планах благоустройства городских парков.<br><br>The mayor told about the plans for the improvement of city parks. | очередная пресс-конференция, планы, программа озеленения города, новые клумбы, красивый город<br><br>another press conference, plans, program of planting greenery in the city, new flowerbeds, beautiful city |
| 9 | В издательстве выходит новая книга известной детской писательницы.<br><br>A new book by a famous children's writer is coming out. | интересный сюжет, история о волшебстве, магические превращения, яркие иллюстрации, первые положительные отзывы, большой тираж<br><br>interesting plot, story about magic, magical transformations, bright illustrations, the first positive reviews, large circulation |
| 10 | Ученые опубликовали результаты масштабного исследования об изменении климата.<br><br>Scientists have published the results of a large-scale study on climate change. | повышение температуры, разрушение озонового слоя, скорость таяния ледников, жизнь на Земле, угроза человечеству<br><br>rise of temperature, ozone depletion, rate of glacier melt, life on Earth, threat to humanity |

Table 1: List of prompts with guide phrases used in experiments

| Generation strategy | $\lambda_0$ | ↓ PPL ± Std | ↓ Rep, % | ↑ SR, % |
|---|---|---|---|---|
| ruGPT3 | – | 8.5 ± 2.7 | 14.53 | 0.00 |
| ruGPT3 + ruRoBERTa | – | 6.9 ± 3.5 | 61.35 | 13.63 |
| ruGPT3 + ruRoBERTa + $\lambda\alpha\Delta$ | 0.1 | 20.6 ± 5.6 | 4.00 | 96.93 |
| | 0.3 | 18.9 ± 4.9 | 2.56 | 99.40 |
| | 0.5 | 18.3 ± 4.0 | 3.62 | 99.67 |

Table 2: Values of quality measures for generating news articles.
Lower values of PPL and Rep correspond to a higher quality model (↓),
higher values of SR correspond to a better model (↑)

The minimum PPL obtained for ruGPT3 + ruRoBERTa is equal to 6.9 (the lower the perplexity, the better the model). However, given the high Re p value for ruGPT3 + ruRoBERTa, we cannot define this strategy as of high quality because the texts contain a large number of repeating fragments. Therefore, without an additional increase of the probability of the first token from the guide phrase, it is impossible to generate good texts. For the ruGPT3 + ruRoBERTa + $\lambda\alpha\Delta$ strategy, the perplexity is greater than for ruGPT3. Such an increase in perplexity compared to the base ruGPT3 model is due to the fact that the shift in the values of the score function (2) in order to guide generation along the given guide phrases is "unnatural" for the model. This causes more "surprise" of the model to the tokens observed in the text.

Figure 5 for the example shown in Figure 3 represents graphs of score functions for ruGPT3 ($score_{ruGPT3}$) and ruRoBERTa ($score_{ruRoBERTa}$) for the top-500 common tokens – candidates for the <mask> token, sorted in descending order of scores in the ruGPT3.

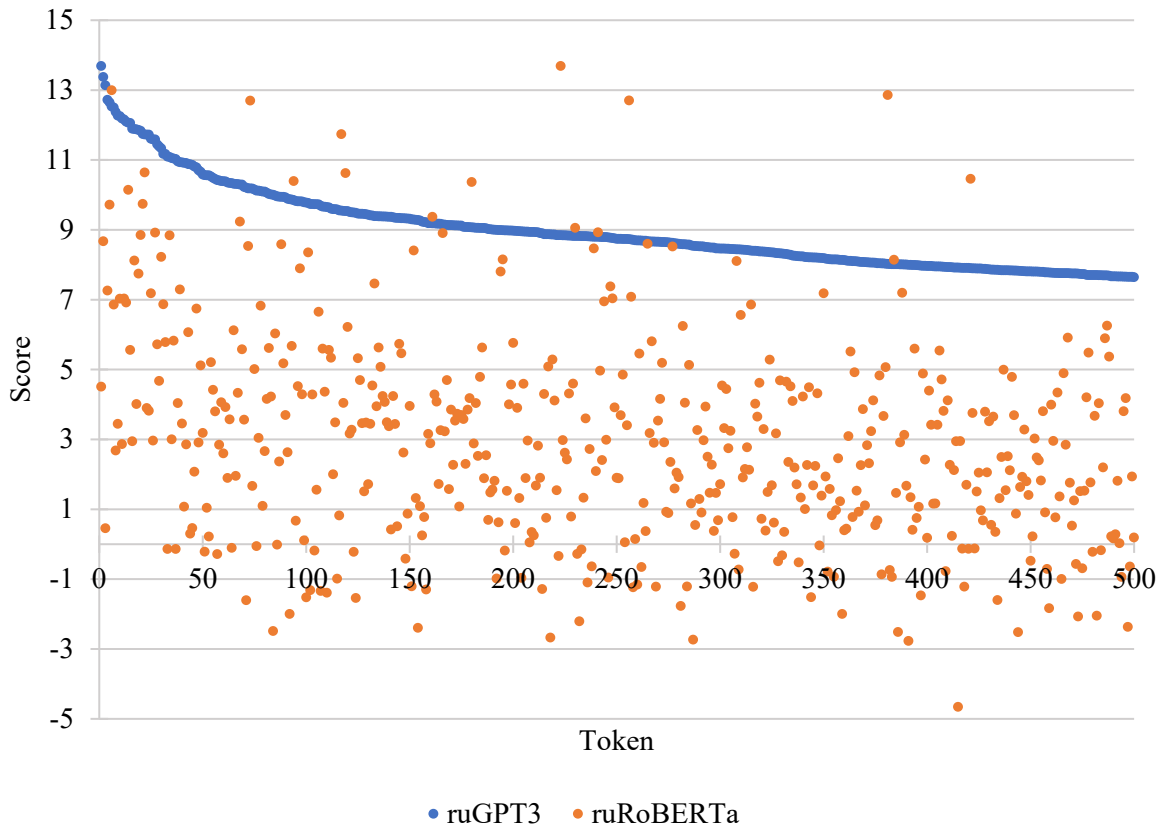

Figure 5: Graphs of the scores for ruGPT3 and ruRoBERTa for the example in Figure 3.
The abscissa shows the serial numbers of tokens in the top-500

Figure 5 shows that with a monotonous decrease of the function $score_{ruGPT3}$ in the graph the values of the function $score_{ruRoBERTa}$ change randomly. The tokens that are more logical for ruGPT3 as those to continue the text based on the left context are not as logical for ruRoBERTa given left and right contexts. Therefore, function (3), which sums the values of the functions $score_{ruGPT3}$ and $score_{ruRoBERTa}$, is a compromise solution for determining the logic of the token at the current generation step.

Table 3 shows the top-10 tokens that are candidates instead of the <mask> token in Figure 3 with the largest values of the score functions for ruGPT3, ruRoBERTa and ruGPT3 + ruRoBERTa.

| ruGPT3 | | ruRoBERTa | | ruGPT3 + ruRoBERTa | |
|---|---|---|---|---|---|
| **Token** | **Score** | **Token** | **Score** | **Token** | **Score** |
| работают | 13.692 | прошла | 13.692 | проходит | 25.526 |
| работает | 13.377 | проходит | 12.995 | пройдет | 22.885 |
| благо | 13.136 | состоялась | 12.857 | прошла | 22.537 |
| уже | 12.720 | состоится | 12.702 | будет | 22.371 |
| идет | 12.648 | пройдет | 12.701 | идет | 22.369 |
| проходит | 12.531 | пройдёт | 11.737 | проводится | 22.210 |
| в | 12.501 | будет | 10.643 | работает | 22.049 |
| действуют | 12.375 | - | 10.625 | планируется | 21.486 |
| более | 12.269 | была | 10.458 | состоится | 21.436 |
| действует | 12.254 | началась | 10.394 | пройдёт | 21.286 |

Table 3: Top-10 tokens with the highest values of score functions for the example in Figure 3

Analyzing Table 3, we can notice that the top-10 ruGPT3 tokens do not reflect the tendency of the narrative in the text to shift towards the guide phrase "очередная пресс-конференция" ("another press conference"). On the contrary, the top-10 ruRoBERTa tokens provide a logical transition to the guide phrase. After summing the functions $score_{ruGPT3}$ and $score_{ruRoBERTa}$ the top-3 tokens include the words "проходит" ("is being held"), "пройдет" ("will be held") и "прошла" ("was held"), which, on average, for both models seem to be the most suitable for both continuing the left context and for a coherent transition to the right context. The result of the method at the $i$-th step for the example in Figure 3 is the next phrase: "Сегодня в городе проходит" ("Today *another press conference* is being held in the city")[6]. In the next step, the <mask> token is moved one position to the right, and the guide phrase is inserted again until it appears in the text. Thus, the proposed method provides a coherent transition to the guide phrase in the text.

Table 4 shows the average computation time for $score_{ruGPT3}$ and $score_{ruRoBERTa}$, as well as the running time of the top-$K$ sampling algorithm in the process of generating one text from the input data from Table 1. The time was calculated when the algorithm was executed on the server with GPU RTX A6000 48 Gb.

Table 5 shows the examples of texts generated using the developed method for different values of $\lambda_0$. Other examples are given in the repository[7].

| **Time of computation** $score_{ruGPT3}$**, sec** | **Time of computation** $score_{ruRoBERTa}$**, sec** | **Time of top-$K$ sampling, sec** | **Total time, sec** |
|---|---|---|---|
| 6.777 | 5.587 | 0.843 | 13.207 |

Table 4: Calculation time of $score_{ruGPT3}$, $score_{ruRoBERTa}$ and the running time of the top-$K$ sampling algorithm when generating one text from the input data from Table 1

---

[6] In Russian, due to specific word order, the phrase "*another press conference*" is followed by "*is being held*".
[7] https://github.com/icecreamz/Collocation2Text.

| $\lambda_0$ | Guide phrases | Generated text |
|---|---|---|
| 0.1 | очередная пресс-конференция, планы, программа озеленения города, новые клумбы, красивый город | Мэр рассказал о планах благоустройства городских парков. Очередная пресс-конференция с журналистами состоялась после того, как депутаты Городской думы приняли решение о выделении в этом году средств для приведения в порядок центральных городских парков и о создании новых зон отдыха и развлечений. - Все планы, которые были озвучены сегодня на заседании Городской думы, были поддержаны депутатами - и это только программа озеленения города. - На площади Ленина появятся новые клумбы и цветники - по инициативе депутатов. - Будет большой и красивый город − |
| | another press conference, plans, program of planting greenery in the city, new flowerbeds, beautiful city | The mayor told about the plans for the improvement of city parks. Another press conference with journalists took place after the deputies of the City Duma decided to allocate funds this year to put the city's central parks in order and create new recreation and entertainment areas. - All the plans that were announced today at the meeting of the City Duma were supported by the deputies - and this is only a program of planting greenery in the city. - New flowerbeds and gardens will appear in Lenin Square - at the deputies' initiative. - There will be a big and beautiful city - |
| 0.3 | интересный сюжет, история о волшебстве, магические превращения, яркие иллюстрации, первые положительные отзывы, большой тираж | В издательстве выходит новая книга известной детской писательницы. Ее интересный сюжет, яркие персонажи и увлекательная интрига никого не оставят равнодушными!Это - настоящая детская история о волшебстве, доброте и волшебных превращениях в мире сказок и магические превращения, происходящие в нашем мире, где все мы являемся частью чего-то большого, прекрасного и удивительного!Очень интересные яркие иллюстрации!И первые положительные отзывы!Это очень не большой тираж!Но это то, что нужно маленьким детям, которым еще не довелось прочитать книгу. |
| | interesting plot, story about magic, magical transformations, bright illustrations, the first positive reviews, large circulation | A new book by a famous children's writer is coming out. Its interesting plot, bright characters and fascinating intrigue will not leave anyone indifferent! This is a real children's story about magic, kindness and magical transformations in the world of fairy tales and magical transformations taking place in our world, where we are all part of something big, beautiful and amazing! Very interesting bright illustrations! And the first positive reviews! This is not a very large circulation! But this is what young children who have not yet had a chance to read a book yet need. |
| 0.5 | новое прочтение, игра актеров, красивые декорации, реакция зрителей, успех | В пятницу в драматическом театре состоялась премьера спектакля. И снова это было новое прочтение пьесы. Игра актеров и режиссера в спектакле не уступает профессионализму постановщиков. Очень понравились красивые декорации, костюмы и реакция зрителей. Первый успех спектакля - это очень хороший старт для молодого театра. А впереди уже новые премьеры. И, конечно, очень интересно наблюдать за тем, как театр развивается. Смотреть фото с концерта можно здесь. |
| | new reading, play of the actors, beautiful scenery, reaction of the audience, success | On Friday, the play premiered at the Drama Theatre. And again it was a new reading of the play. The play of the actors and the director in the performance is not inferior to the professionalism of the directors. I really liked the beautiful scenery, costumes and the reaction of the audience. The first success of the performance is a very good start for the young theatre. And there are new premieres ahead. And, of course, it is very interesting to watch how the theater develops. You can see photos from the concert here. |

Table 5: Examples of texts generated by the proposed method. The prompts are highlighted in blue, the guide phrase – in red. The spelling of the generated texts is given using the original variant of the model

## 6  Conclusion

The proposed *Collocation2Text* method allows generating texts according to the user-specified list of guide phrases that represent the text storyline. Guide phrases can be single words and collocations. The method combines the advantages of autoregressive and autoencoding language models. The first model can well predict the next token in the sequence by the previous context, and the second model – the missing word in the text. We applied this method to the Russian language. The experiments on the news articles generation showed that the method ensures the occurrence of a high proportion of guide phrases in the text (more than 96% in SR), provides a low repetition of the text (less than 4% in Rep), which is even less than the repetition of the basic generative language model. However, this increases the text perplexity. The analysis of the tokens with the highest value of the score function showed that the method contributes to a coherent transition between the previously generated text and the guide phrases that are supposed to occur in the text.

## Ethical Considerations

Information reliability in automatically generated texts is determined by the information reliability used to pre-train the models. It is known that large language models, including ruGPT-3, are capable of generating texts that cannot be distinguished from human-written texts. The method proposed by us can generate texts containing false and unreliable information. The use of guide phrases that contain factually incorrect content will result in the generation of incorrect or false information that can be used to harm, for example, to spread personalized misinformation. Our intention is to make social communications easier. We hope that this method will be used in socially positive applications. Providing open access to such methods will help to develop ways to detect them.

## Acknowledgements

This work was supported by Russian Science Foundation, project № 22-21-00885, https://rscf.ru/en/project/22-21-00885.